# Text Windows and Phrases Differing by Discipline, Location in Document, and Syntactic Structure[*]

Robert M. Losee
Manning Hall, CB #3360
U. of North Carolina
Chapel Hill, NC, 27599-3360
USA

losee@ils.unc.edu

**Abstract**

Knowledge of window style, content, location and grammatical structure may be used to classify documents as originating within a particular discipline or may be used to place a document on a theory versus practice spectrum. This distinction is also studied here using the type-token ratio to differentiate between sublanguages. The statistical significance of windows is computed, based on the the presence of terms in titles, abstracts, citations, and section headers, as well as binary independent (BI) and inverse document frequency (IDF) weightings. The characteristics of windows are studied by examining their within window density (WWD) and the $\mathcal{S}$ concentration (SC), the concentration of terms from various document fields (e.g. title, abstract) in the fulltext. The rate of window occurrences from the beginning to the end of document fulltext differs between academic fields. Different syntactic structures in sublanguages are examined, and their use is considered for discriminating between specific academic disciplines and, more generally, between theory versus practice or knowledge versus applications oriented documents.

---

[*]The author wishes to thank Kent Parks for his diligence compiling several of the databases.

# 1 Introduction

The nature of term groupings, phrases, and text windows in documents is not fully understood, yet the importance of term clusters is obvious to those in disciplines who study text. While models and implementations of retrieval, filtering, and indexing systems often address terms individually, knowledge about multi-term structures should enhance the quality of systems based on more accurate models of language and feature relationships. For example, increased knowledge about the relationships between features may lead to improved performance in retrieval systems and in the automatic classification of documents. We examine below several characteristics of phrases and text windows, including their number, location in documents, and grammatical construction, in addition to studying variations in these window characteristics across disciplines. We examine some of the "linguistic regularities" (Sager, 1981) for individual disciplines, and suggest families of regularities that may prove helpful for the automatic classification of documents, as well as for information retrieval and filtering applications.

## Disciplines

Academic disciplines have different objects of study, methods, and philosophical attitudes (Cole, 1983, 1978; Cole, Cole, & Dietrich, 1978; Lodahl & Gordon, 1972; Price, 1986). The language used by authors in these areas may differ because of the nature of the material presented, as well as the variations in discipline specific language grammars, vocabularies, and literary styles (Bonzi, 1990, 1984; Damerau, 1993; Haas & He, 1993; Losee & Haas, 1995; Sager, 1981). Learning the nature of consistent differences in these sublanguages may allow one to determine the sublanguage being used in a document of unknown disciplinary origin.

Academic fields may be placed within a spectrum of disciplines or in groupings based on a variety of characteristics. A common distinction is to study disciplines on a spectrum from "hard" to "soft" sciences (Cole, 1983; Lodahl & Gordon, 1972; Losee & Haas, 1995; Price, 1986). Similarly, disciplines may be characterized as "donor" or "borrower" disciplines (Narin, Carpeter, & Berlt, 1972; Losee, 1995b).

Research between and within disciplines may be seen as ranging from "theory" to "practice," with some disciplines or subdisciplines judged as more theoretical while others are very applications oriented, as is often found in professional and engineering disciplines. Within a discipline such as physics, for example, published research on string theory and mathematical physics is clearly more theoretical than articles on instrumentation or experimental physics. Between disciplines, we find areas that are more academic (e.g. literature) as opposed to those that are more applied, such as mechanical engineering or professional fields, although most fields (such as Literature) have applied aspects (e.g. teaching composition) as well as theoretical subdisciplines within these broader fields. The ability to identify theoretical or practical articles may be of benefit to those performing searches for documents or by individuals or systems classifying documents. We will examine whether writings that are more theoretical exhibit a greater rate of term repetition, possibly because they are more specific and work within smaller domains in which the language may not be as rich.

## Windows and Phrases

While a full understanding of the syntax, semantics, and phonology of natural language is the long term goal of many scholars (Phillips, 1985; Newmeyer, 1986), an intermediate goal may be the understanding of the nature of smaller "chunks" of natural language, often consisting of a few terms (Craven, 1989; Lauriston, 1994; Maeda, 1981; Milic, 1991). Groups of terms are also of interest if one assumes that humans may process language in these small chunks, suggesting a focus on phrases as a natural unit of syntactic processing. The importance of studying phrases is emphasized by Strzalowski (1995) who argues that, for retrieval systems, "the use of phrasal terms is not just desirable, it becomes necessary." Windows and phrases are not just a function of a single language, but are found in a variety of languages (Jacquemin, 1995).

Term occurrences within natural languages may be modeled statistically (Charniak, 1993), and statistical methods may be used in specific applications, such as in parsing (Weischedel, Schwartz, Palmucci, Meteer, & Ramshaw, 1993). For example, the terms *cystic* and *fibrosis*, are very often found in adjacent positions in the CF database, occurring separately far less frequently, suggesting that the two terms may usefully be treated as a single unit for parsing or retrieval purposes. Similar co-occurrences may be studied qualitatively (Caroli, 1995).

A *window* within natural language text is a set of $\omega$ consecutively occurring terms, where $\omega$ is referred to as the *window size*. Windows are of interest when they are expected to have a disproportionate number of terms of interest for a particular application, e.g., retrieval or automatic classification. If we limit our study to those windows that are *special* in some way, we expect that the availability and use of the term clusters of interest in windows will result in improved performance discrimination between classes of documents when knowledge of windows is explicitly included in the design of a discriminating system such as a retrieval, classification, or filtering system.

The size or span of text windows has been an actively studied by scholars (Haas & Losee, 1994; Haas & He, 1993; Losee, 1994; Martin, Al, & van Sterkenburg, 1983; Smadja, 1993). Information retrieval systems may serve as a testbed for the analysis of term grouping characteristics by studying retrieval performance assuming differently sized windows or different grammatical structures. The sizes of windows of potential interest also may arise from the study of phrases that appear interesting (e.g., noun phrases, sentence subjects, clauses, etc) and incidentally happen to have a certain size. More pragmatic concerns have driven others to compare retrieval system performance under different conditions, noting those changes in window sizes that result in particular levels or changes in retrieval system performance. Phrases may be useful in indexing documents (Jones, Gassle, & Radhakrishnan, 1990) and in the direct retrieval of documents (Fagan, 1989). Groups of terms have been identified as related and treated as a unit through the explicit study of the statistical dependence existing between terms (Chow & Liu, 1968; Croft, 1986; Lam & Yu, 1982; Losee, 1994; Van Rijsbergen, 1977; Yu, Buckley, Lam, & Salton, 1983). The experimental performance of retrieval systems that use terms clustered into windows has been studied (Akers, 1995; Haas & He, 1993; Haas & Losee, 1994; Losee, 1994). Performance of retrieval systems using groups of terms that are statistically related may be computed analytically rather than experimentally (Losee, 1995a, 1996a).

This study will examine several non-retrieval based aspects of windows. Documents may be classified by using any of a number of available document characteristics. While there are certainly features that discriminate between documents on theory and practice other than those presented here, we are able to show a set of distinct grammatical and structural difference between these areas that may be useful in automatic classification of documents. This, combined with vocabulary based methods (Losee & Haas, 1995), may produce fast and accurate classification systems or may be used in identifying key indexing or sublanguage terms and phrases. Other techniques useful for automatic classification and labeling documents for future retrieval will be examined below.

## 2  Statistically Significant Windows

All contiguous sequences of terms in an article are intended to have some meaning or semantic value. In this work, we move beyond treating all text windows or groups of terms as being equally important, to identifying those windows that are statistically unlikely to contain as many *special* (or "$\mathcal{S}$") terms as they do. Terms are labeled as $\mathcal{S}$ terms for purposes here if they are in a certain part or field of a document (e.g. title or abstract), occur in a subject dictionary, or have a term weight that exceeds an arbitrarily selected cutoff value.

A cluster of terms may be treated as either statistically significant or as the result of the random clustering of terms (Dunning, 1993). Determining that a window is *special* may be done statistically by designating as special only those windows that are statistically very unlikely to have as many *special* terms as they do have, assuming that the $\mathcal{S}$ terms are randomly distributed. The significance of a window with $r$ $\mathcal{S}$ terms is determined probabilistically based upon the probability that the $r$ $\mathcal{S}$ terms would occur in a window of size $\omega$. While defining a window in terms of the presence of an $\mathcal{S}$ term will force most small windows containing an $\mathcal{S}$ term to be statistically significant, studying the significance of windows is effective in looking at the clustering of $\mathcal{S}$ terms. These *statistically significant windows* ($\mathcal{SW}$s) are identified automatically, making this method suitable for use in automated document retrieval, filtering, classification, and indexing systems.

Different methods of defining the set of $\mathcal{S}$ terms are used below in the study of windows and their significance. Once the set of $\mathcal{S}$ terms has been defined, the significance of a window may be statistically determined by computing the probability that there are exactly $r$ terms in a window of size $\omega$, as

$$B(r; \omega, p) = \binom{\omega}{r} p^r (1-p)^{\omega - r}, \qquad (1)$$

where $p$ is the probability that a term token in the text is in the set of $\mathcal{S}$ terms. We use the cumulative binomial distribution, and for purposes here, a window was decided to be significant if the probability that a window would have as many $\mathcal{S}$ terms as it has was less than 3%. This was an arbitrarily chosen value and allows us to assert that there is at least a 97% chance that a randomly selected window would have fewer than this many $\mathcal{S}$ terms.

While the set of $\mathcal{S}$ terms may be defined as all those terms in a particular document component and are thus special because they occur in subject bearing parts of documents such as the title or the set of section headers, a set of $\mathcal{S}$ terms also may be defined as the set of terms having a particular quantitative characteristic. We may compute the weight for a term, assuming the binary independence model, as

$$W_{BI} = \log\left(\frac{p/(1-p)}{q/(1-q)}\right) \quad (2)$$

where $p$ is the probability that a document in the given database has the term and $q$ represents the probability that a given document in the combined set of databases has the term. For example, the presence of the terms *cystic, fibrosis,* and *were* are positive indicators that a document is from the database of documents on cystic fibrosis (from among the databases examined here). The terms *cystic* and *fibrosis* are easily explained and expected in this database, while the term *were* probably occurs in this database with greater relative frequency due to the more frequent use of the past tense in medical documents, where symptoms and treatments are described, than in some other disciplines studied, such as physics.

The Inverse Document Frequency (IDF) weighting is a popular term weighting system that is a special case of the BI weighting system (Losee, 1988). It is computed as

$$W_{IDF} = -\log q. \quad (3)$$

essentially the log of the proportion of documents with the term.

## 3   Some Hypotheses and Assumptions

We make several hypotheses in our research. We broadly assume that statistically significant windows ($\mathcal{SW}$s) are not randomly distributed throughout the fulltext of documents. The first hypothesis, the *fulltext sublanguage differentiation* hypothesis, suggests that

> there are vocabulary, grammatical, and stylistic differences between the fulltext of documents in different disciplines. Terms in document components are distributed differently throughout the fulltext documents written in different sublanguages.

If this is true, this variation may be used to identify, or to assist in the identification of different disciplinary writing styles and thus identify the disciplinary source of a document.

A second hypothesis, the *theory-practice term density* hypothesis, suggests that terms are distributed differently in the fulltext of theory based documents than in practice based documents. Authors writing in the professional, engineering, and applied disciplines reuse terms at a lower rate than in the more theoretical liberal arts and sciences which have more specialized vocabularies.

This may be due to the greater depth of the theoretical literature and the concurrent limitation placed on the vocabulary in these specialized disciplines. Differences in term frequencies, as evidenced by a difference in the type-token ratios for fragments of text for different document styles, have been studied by Tagliacozzo (1976). If our hypothesis is true, theoretical and applied documents may be identified based on the density of the terminology. Further, studying sublanguage type-token ratios may allow the distribution of disciplinary vocabulary to be modeled using Zipf's law, with different parameter values for different academic fields (Sho Chen & Leimkuhler, 1989).

Two sublanguage window assumptions made in our study of windows that are statistically likely to be more than a collection of randomly selected terms are that:

1. terms that are $\mathcal{S}$ terms are most likely to be in an $\mathcal{SW}$ .

2. $\mathcal{SW}$s are most likely to contain $\mathcal{S}$ terms.

These assumptions are almost tautological in our environment.

## 4  Experimental Methods

### Fulltext Databases

Four fulltext databases were used in this project. These databases represent documents written in different disciplinary contexts. The CF fulltext database contains the full text of the first 123 fulltext articles selected from a larger database of fulltext articles (Moon, 1993) extracted and developed from a database of 1239 document representations indexed by the subject heading *Cystic Fibrosis* in the MEDLINE system (Shaw, Wood, Wood, & Tibbo, 1991). Almost all of these documents have a title, abstract, fulltext, and a set of major Medical Subject Headings (MESH).

An unnamed machine readable medical dictionary was obtained from the PC-SIG library of public domain software (Disk 4160, 13th edition, CDROM version) and supplemented to include most of the specialized terms found in the CF database. The terms found in this dictionary are considered to be medical sublanguage terms for purposes here.

Several subject databases were extracted from the sets of technical reports and preprints of articles to be published maintained online by the Lawrence Livermore Laboratory systems group. The fulltext of technical reports were downloaded from this system and edited to produce the desired data format. Titles, abstracts, fulltext, citations and section headers were marked and these terms used as *special* terms for the study below. Beginning with recent articles, documents were downloaded and those with citations and those appearing to be in a relatively standard form of LaTeX, the standard document format for the LLL system, and could be stripped using the standard "detex -l" program and option, were selected for inclusion in our study. Use of the detex program introduces some noise into the process, such as when there are nested commands in conjunction with special characters or symbols, or when the symbols may be separated when they should be together. These errors occur only occasionally but would have some impact on the accuracy of the grammatical analysis obtained from these portions of documents. Documents from LLL were selected slightly differently for each of the disciplinary databases, with differences being due to a number of factors, including when the fulltext archives were started (some are quite recent), the frequency with which documents were posted to these archives, and the number of documents posted in a standard form of LaTeX (most archived documents, but not all, were in such a form). All documents were from the 1994 to 1995 period.

The linguistics database (referred to as CMPLING) consists of 67 fulltext documents from the LLL computational linguistics archive. There are 70 fulltext documents in the High Energy Physics-Theoretical database (HEPTH) and 75 documents in the High Energy Physics-Experimental database (HEPEX)

The four databases (CF, HEPEX, HEPTH, and CMPLING) are the primary databases used in our study. The CMPLING, HEPTH, and HEPEX, databases do not have MESH subject headers, and the CF database has neither section headers nor citations marked.

Table 1: Average number of term tokens and term types in different fields.

|  | Average Number of Tokens | | | | Average Number of Unique Types | | | |
|---|---:|---:|---:|---:|---:|---:|---:|---:|
|  | HEPTH | HEPEX | CMPLING | CF | HEPTH | HEPEX | CMPLING | CF |
| Fulltext | 2858.9 | 2529.4 | 4610.1 | 2094.93 | 597.1 | 596.5 | 892.2 | 572.9 |
| Title | 8.2 | 10.0 | 7.1 | 12.4 | 8.1 | 9.7 | 7.0 | 11.6 |
| Abstract | 90.4 | 83.0 | 97.5 | 126.0 | 57.9 | 54.3 | 65.0 | 71.8 |
| Citations | 232.0 | 233.5 | 373.8 |  | 103.1 | 97.7 | 182.8 |  |
| Sec. Hdrs. | 42.2 | 20.9 | 29.6 |  | 19.5 | 16.0 | 22.3 |  |
| Major MESH |  |  |  | 5.2 |  |  |  | 5.1 |
| Minor MESH |  |  |  | 19.4 |  |  |  | 18.1 |
| BI in fulltext | 184.7 | 128.0 | 353.1 | 84.8 | 12.8 | 14.6 | 23.7 | 6.1 |
| IDF in fulltext | 573.0 | 490.1 | 1428.3 | 642.0 | 245.1 | 237.2 | 466.4 | 284.1 |

The analysis of window characteristics was accomplished thorough custom-written programs. These Bourne shell scripts and gawk programs were executed on a Unix workstation.

## Tagging Parts of Speech

Most information retrieval systems use little or no information that is provided by the parts-of-speech of terms in the document, although this information may be included and may improve retrieval performance (Burgin & Dillon, 1992; Lankhorst, 1995; Losee, 1996b; Yang, 1993). For example, a document with the sentence "dog bites boy" places "boy" in a different role than the sentence "boy bites dog." Knowledge of the grammatical parts-of-speech of the term "boy" in the documents of a database might help one retrieve the document containing boy as biter or as bitten, depending on the needs of the searcher.

Terms in out fulltext documents were labeled with their parts-of-speech as assigned by the Brill tagger (Brill, 1994), but other tagging procedures could have been used (Weischedel et al., 1993). The Brill tagger assigns part-of-speech tags consistent with the tagset used with the Penn Treebank. While the tagger may be trained to learn to assign the parts-of-speech more accurately for a particular database, this option was not used. Much of this study therefore may be replicated using the Brill tagger as available during the summer of 1995.

## The Location of Significant Windows in Documents

The number of $\mathcal{SW}$s per 100 windows and the relative location of the center of the windows in the fulltext of the document are displayed graphically below. Locations are in the fulltext of document, excluding other fields from the database in question. Thus, the beginning of the fulltext is after the abstract and before the citations, and does not include these or other features, i.e., section headers.

The statistical significance of these results may be studied by performing separate analyses with the odd numbered and even numbered documents from the CMPLING database. Each half thus contains about 33 documents. The results shown in Figure 1, being from smaller half databases, are more likely to show variance than the results shown in other analyses below that use a full database, but allow us to gauge the variation that might occur. In all cases, the even and odd databases had the same rough shapes for the curves for CMPLING, with minor variations. The two end points of the graphs use half the data of the other points in the graph, allowing us to examine these special points in the documents; the number of windows at these points is more susceptible to noise than the interior points. Use of the fulltext databases give results that show some variation, but which are likely to be within 10% of the field values most of the time. This degree of accuracy is satisfactory for our purposes, since the variation still allows us to see trends reflective of the disciplines and also allows us to distinguish between the disciplines based on the characteristics of the $\mathcal{SW}$s.

It should be noted when analyzing these window location graphs that each data point is derived separately from other data points on the same line. A graphed line for discipline $x$ that is always above a line for discipline $y$, showing a series of $n$ data points for each discipline and at *each* of these $n$ points, the value for $x$ exceeds that for the corresponding point for $y$. Each point represents the average number of $\mathcal{SW}$s that were found in a small (6%) portion of the fulltext of the document, with the average taken over thousands of possible locations.

Figure 1: The statistically significant windows, from the beginning of fulltext to the end, for even and odd numbered documents in the CMPLING database. Windows in even and odd number documents are graphed separately to provide some indication of the variance found. The "A" represents windows derived from $\mathcal{S}$ terms from abstracts and "C" as windows derived from $\mathcal{S}$ terms being those terms in citations.

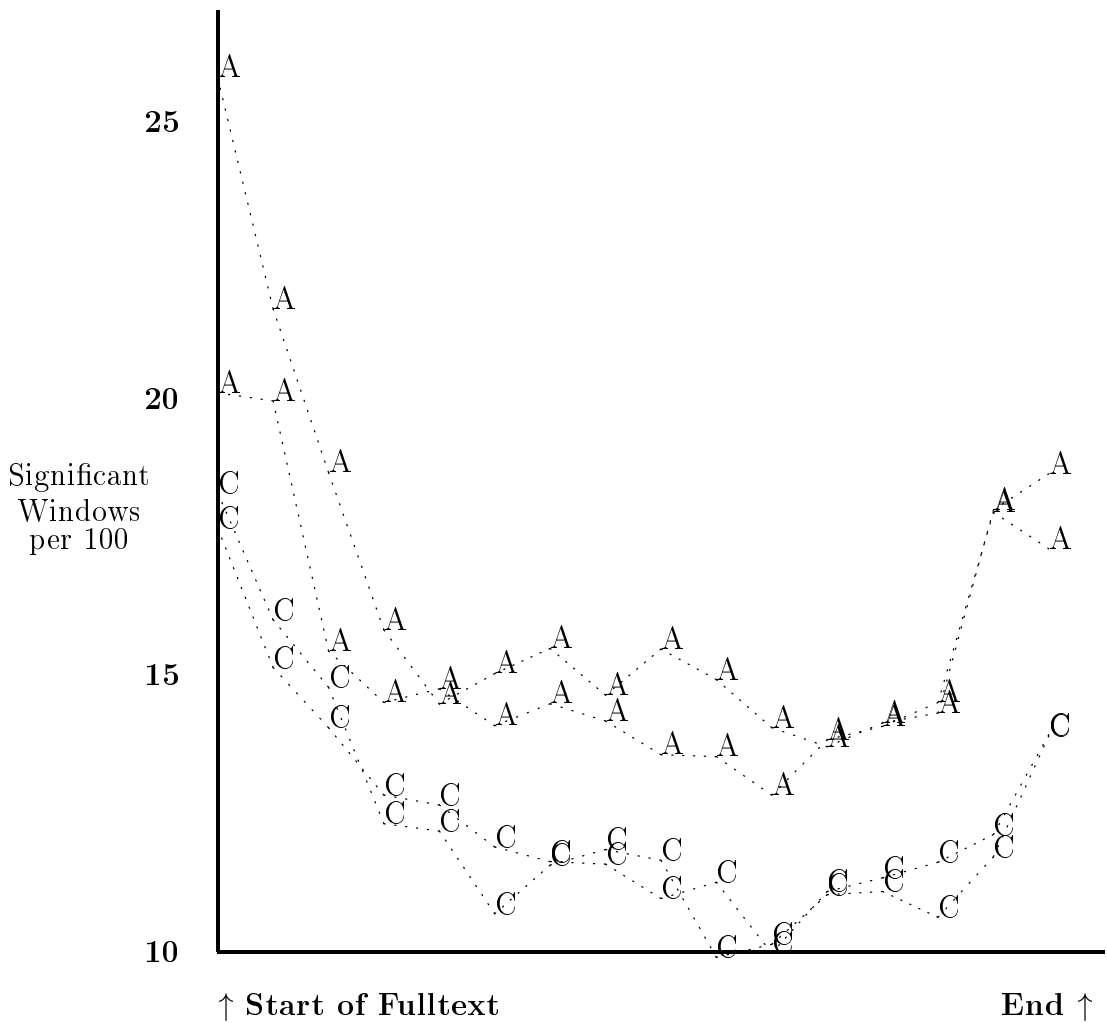

## 5  Measuring Characteristics of $\mathcal{S}$ Terms & Windows

There are numerous ways that we can understand and measure the tendency of $\mathcal{S}$ terms to cluster together into windows. Using different measures of diversity allows us to study spread or clustering consistent with a variety of assumptions (Losee, 1990). Studying the degree to which windows produced by the methods proposed here differ from windows composed of randomly distributed $\mathcal{S}$ terms, for example, allows one to measure the degree to which the observed clustering provided by a set of windows improves clustering of $\mathcal{S}$ terms beyond the expected level of clustering.

The results described here use a measure of $\mathcal{S}$ windowing quality, $\mathcal{Q}$, that combines two other, easily interpretable quantities, both of which should be maximized for our applications. Text windows should bring together terms and concepts of interest if the windows are to be useful. We may measure the degree to which these special terms are concentrated in $\mathcal{SW}$s by measuring the probability that a term is in an $\mathcal{SW}$ given that it is an $\mathcal{S}$ term, $\Pr(s \in \mathcal{SW} | s \in \mathcal{S})$, where we are measuring the probability that term $s$ is in the set of $\mathcal{SW}$s given that $s$ is in the set of $\mathcal{S}$ terms. This value is referred to as $\mathcal{S}$ *Concentration* (SC); it reaches the maximum of 1 when all the $\mathcal{S}$ terms are in significant windows and approaches 0 when none of the terms in the $\mathcal{SW}$s are $\mathcal{S}$ terms. We may measure the *Within Window Density* (WWD) as the probability that $s$ is in $\mathcal{S}$ given that $s$ is in an $\mathcal{SW}$, $\Pr(s \in \mathcal{S} | s \in \mathcal{SW})$. This probability approaches 1 when all the terms in a window are $\mathcal{S}$ terms, and approaches 0 when none of the terms in $\mathcal{SW}$s are $\mathcal{S}$ terms.

The $\mathcal{Q}$ measure, a combination of the $\mathcal{S}$ concentration and Within Window Density measures, may be computed as the product of the two measures, reaching 1 when both are 1 and approaching 0 when both individual measures approach 0. We compute

$$\mathcal{Q} = \Pr(s \in \mathcal{SW} | s \in \mathcal{S})^\alpha \Pr(s \in \mathcal{S} | s \in \mathcal{SW})^\beta \qquad (4)$$

where $\alpha$ and $\beta$ are weights for the SC and WWD measures. For our purposes, we set $\alpha = \beta = 1$, making $\mathcal{Q}$ simply the product of the within window density and the $\mathcal{S}$ concentrations. Some window characteristics for our databases are given in Table 2. Windows of size three are used, with the $\mathcal{S}$ terms drawn from those terms in the fields indicated on the left. These results, as well as those for Table 3, which shows similar data with stop words excluded from the sets of $\mathcal{S}$ terms, are discussed in the sections below.

Figure 2: The distribution of CF fulltext significant windows from the beginnings of documents to their end. Fields used for producing $\mathcal{S}$ terms are represented as Abstracts (A), Title (T), and Dictionary (D).

↑ **Start of Fulltext**            **End** ↑

Table 2: $\mathcal{S}$ Concentration (SC), the probability that an $\mathcal{S}$ term is in an $\mathcal{SW}$, Within Window Density (WWD), the probability that a term is an $\mathcal{S}$ term given that it is in an $\mathcal{SW}$, and $\mathcal{Q}$, the product of $WWD$ and $SC$ for fulltext windows, where $\mathcal{S}$ terms are from the fields on the left for the four databases shown across the top. Windows are of size 3.

|  | *HEPTH* | | | *HEPEX* | | | *CMPLING* | | | *CF* | | |
| --- | --- | --- | --- | --- | --- | --- | --- | --- | --- | --- | --- | --- |
| *From:* | SC | WWD | $\mathcal{Q}$ | SC | WWD | $\mathcal{Q}$ | SC | WWD | $\mathcal{Q}$ | SC | WWD | $\mathcal{Q}$ |
| *title* | .69 | .34 | .23 | .59 | .29 | .17 | .87 | .41 | .36 | .56 | .29 | .16 |
| *abstract* | .26 | .15 | .04 | .24 | .14 | .03 | .32 | .19 | .06 | .34 | .21 | .07 |
| *citation* | .39 | .22 | .09 | .35 | .17 | .06 | .36 | .20 | .07 | | | |
| *sec. hdr.* | .51 | .28 | .15 | .63 | .32 | .20 | .60 | .30 | .18 | | | |
| *major MESH* | | | | | | | | | | .98 | .54 | .53 |
| *dictionary* | | | | | | | | | | .42 | .23 | .10 |
| *BI* | .98 | .43 | .42 | .99 | .45 | .45 | .95 | .47 | .45 | .92 | .43 | .43 |
| *IDF* | .38 | .18 | .07 | .38 | .19 | .07 | .37 | .22 | .08 | .36 | .20 | .08 |

Table 3: Using fields with stopwords removed, Within Window Density, $\mathcal{S}$ Concentration, and $\mathcal{Q}$, are computed from the four fields on the left for the four databases shown across the top. Windows are of size 3.

|  | HEPTH | | | HEPEX | | | CMPLING | | | CF | | |
|---|---|---|---|---|---|---|---|---|---|---|---|---|
| From: | SC | WWD | $\mathcal{Q}$ | SC | WWD | $\mathcal{Q}$ | SC | WWD | $\mathcal{Q}$ | SC | WWD | $\mathcal{Q}$ |
| title | .99 | .51 | .51 | .99 | .52 | .51 | .99 | .55 | .55 | .96 | .49 | .47 |
| abstract | .75 | .41 | .31 | .73 | .40 | .29 | .85 | .45 | .38 | .55 | .31 | .17 |
| citation | .97 | .49 | .47 | .94 | .52 | .49 | .78 | .41 | .32 | | | |
| sec. hdr. | .96 | .49 | .47 | .98 | .49 | .48 | .99 | .50 | .50 | | | |

# 6 Distinguishing Characteristics of the Disciplinary Databases

Significant windows ($\mathcal{SW}$s) are spread unevenly throughout the fulltext of documents. For example, $\mathcal{SW}$s containing special ($\mathcal{S}$) terms from documents' titles are distributed differently in fulltext from $\mathcal{SW}$s with $\mathcal{S}$ terms from the abstracts, and these variations, in turn, differ between databases and disciplines. Knowledge of these differences may provide assistance in determining the topic and experimental or research orientation of documents.

Disciplines are the source of many of the differences discussed below, but other factors also may produce variation. One cause for differences may be the *typographic effect*. Documents in CMPLING, HEPTH, and HEPEX have been stripped of their markup and formating information in our study with the standard "detex" program that removes LaTeX constructs. The stripping process often involves arbitrary decisions. For example, should the formating for the expression $n^2$ be decomposed into one symbol complex, e.g. $n2$, or should the exponent be separated, e.g. "$n$ 2"? The decision to take one approach instead of another affects the number of terms calculated to be present and controls the "vocabulary" in the $\mathcal{S}$ fields.

Term type and token frequencies have an impact on the relative frequencies of $\mathcal{SW}$s in fulltext. The WWD, SC, and $\mathcal{Q}$ figures in Table 2 are partially determined by the average term frequencies within the fields in question. For example, the data for $\mathcal{S}$ terms derived from titles in Table 2 is ordered across the table in a manner similar to the average number of types and tokens for each of the corresponding databases in Table 1. The strongest relationships exist between the average number of types and the $\mathcal{S}$ Concentration (SC). While this relationship holds for titles, abstracts, and BI, it does not hold for citations, section headers, or IDF. Clearly, more is involved in windowing behavior than can be explained with only knowledge of token and type frequencies for the $\mathcal{S}$ term producing fields.

## Specific Disciplines and Fields

Disciplines such as physics may be characterized by window characteristics. Table 2 shows that the technical reports in physics have lower SC, WWD, and $\mathcal{Q}$ values than is found in the other fields tested. This is largely due to the lower average number of tokens and types in abstracts, as shown in Table 1.

The distribution of $\mathcal{SW}$s in CF fulltext documents may be used to separate these documents from documents originating in other disciplines. Figure 2 shows the rate of occurrence of $\mathcal{SW}$s in the CF fulltext databases, where the $\mathcal{SW}$s are based on sets of $\mathcal{S}$ terms derived from different fields. The windows whose data points are shown on the graphs are all of length 2. This database is distinguished best from the other databases by the rise in the rate of $\mathcal{SW}$ occurrences once we move past the first quarter of the document. Unlike the usual shape of the abstract curve (the split-half or even-odd data shown in Figure 1 for CMPLING is typical), authors of articles in the medical literature use the terms in the abstract more frequently in the body of the text and with more regularity than is found in other disciplines. This may be due both to a consistent and unambiguous vocabulary for medical discussions as well as to a greater degree of focus in literature describing a specific medical condition and its treatment. If this is correct, we would expect $\mathcal{SW}$s from the engineering and possibly experimental literature to show the same shape as $\mathcal{SW}$s composed of $\mathcal{S}$ terms taken from the abstract.

The data represented by Figure 3 show how the $\mathcal{SW}$s composed of terms from a document's abstract varies from one discipline to another. We note that for the second half of the documents (the right half of the graph) the more practical the field's orientation, the higher the rate of occurrence of $\mathcal{SW}$s. All four disciplines start near each other on the left and all but CF end near each other.

We find the same phenomenon occurring with the three databases for which we have citation data, as shown in Figure 4, where $\mathcal{S}$ terms are terms found anywhere in citations for the document. The HEPEX literature is higher for all points except for the initial one. Terms in the citation should act like title terms except that there should be a greater variety of terms in the citations; this relationship is discussed more fully by Kwok (1975).

The location of $\mathcal{SW}$s for $\mathcal{S}$ terms occurring in a subject dictionary are shown in Figure 2. The windows with dictionary terms occur with their highest rate starting about one quarter of the way through the fulltext and then appear to decrease in frequency as one moves through the document. These discipline specific terms and phrases may be used with greater frequency when the problem is being defined and the general literature is discussed. These parts of an article give way to more serious discussion about the problems and questions the researcher faces, with the use of specialized terminology slowly decreasing as less subject-specific terms are used, results are explained, and conclusions are drawn.

Sets of $\mathcal{S}$ terms may be formed from those terms having a computed numeric weight exceeding an arbitrary cutoff. Terms with a binary independence weight $W_{BI} >= 5$ or a term with an inverse document frequency value $W_{IDF} >= 4$ are treated as $\mathcal{S}$ terms here. The choice of these cutoffs determines what terms are $\mathcal{S}$ terms, and thus the values for the data below and in Tables 1 and 2 are completely dependent on the arbitrary choice of the cutoff values. No useful patterns are visible with graphs produced using the BI and IDF based $\mathcal{S}$ terms.

## 7  Theory versus Practice

The increased number of $\mathcal{SW}$s per 100 windows for the more applied literature may be due to the decreased variance of terminology for the theoretical literature which may exhibit a greater gap and correspondingly less overlap between the terms present in the various fields discussed here and the fulltext. This supports the *theory-practice term density hypothesis.*

Table 1 shows limited data differences between documents stressing theory and documents stressing practice. Looking at the relationship between HEPTH and HEPEX, we find that the theoretical documents have higher numbers of types of abstracts, citations, section headers, and IDF derived $\mathcal{S}$ terms. Title terms have lower average type and token frequencies for HEPTH and for CMPLING, while the two experimental databases have higher average numbers of types. This difference is accentuated if one examines the average number of tokens. The average number of fulltext and BI tokens is higher for the theoretical disciplines than for the practical disciplines. Other variations between theory and practice may be seen in the citations and section headers, which show the ordering CMPLING, HEPTH, HEPEX. Because these fields are not indexed for CF, we hesitate to make more than weak claims for the theory versus practice distinction for citations and section headers.

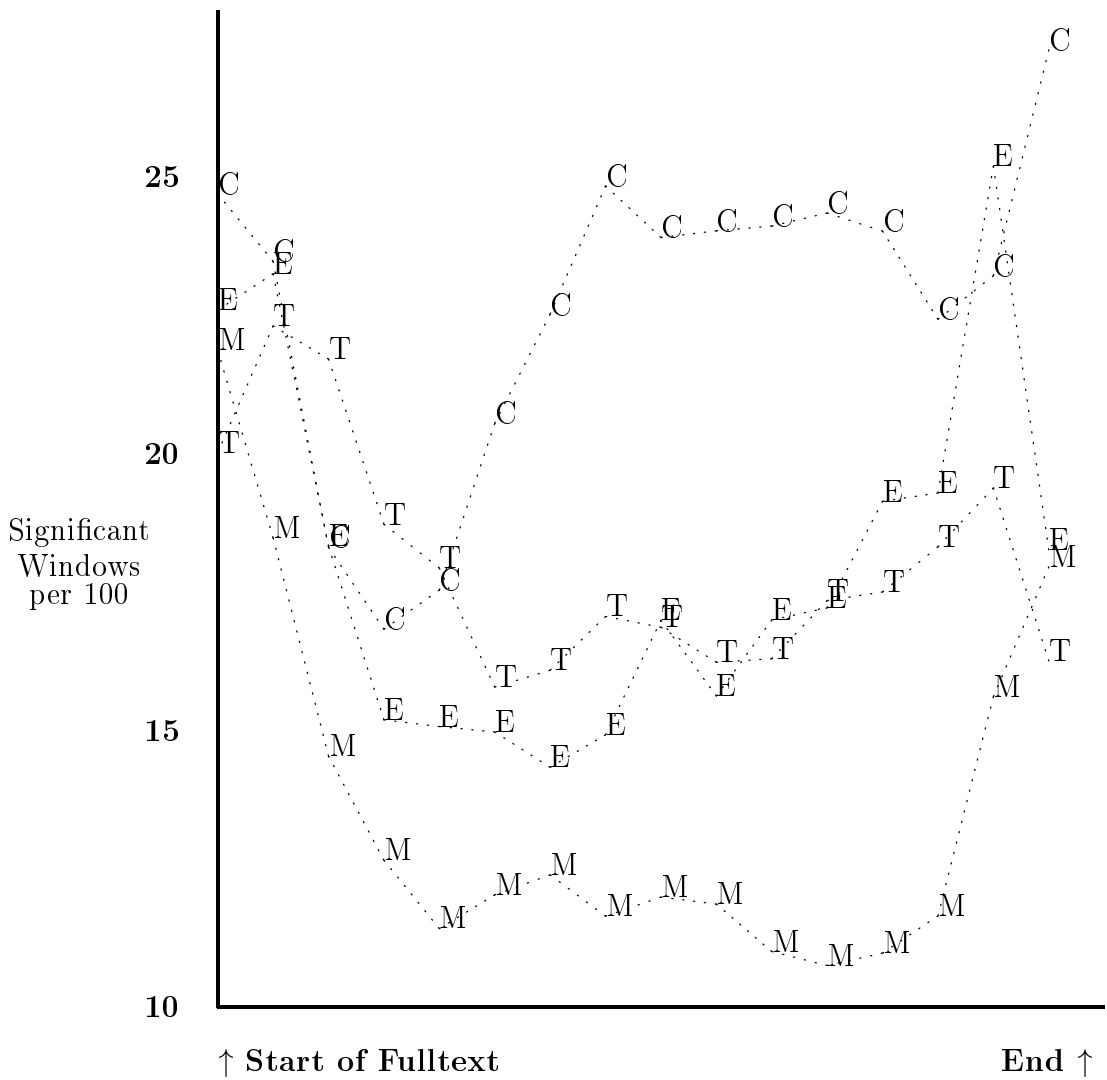

Figure 3: Abstracts may indicate the shift from the professional and experimental literature toward the more theoretical literature. Databases used are CF (C), HEPTH (T), HEPEX (E), and CMPLING (M).

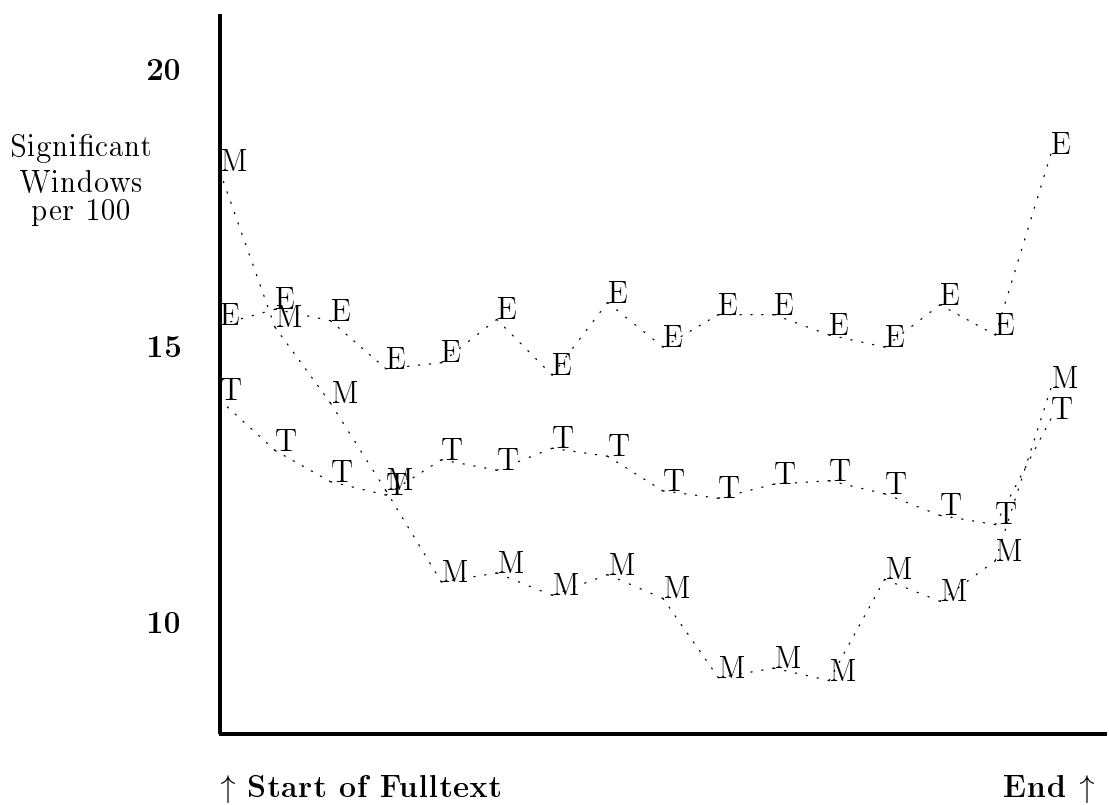

Figure 4: The distribution of significant windows with $\mathcal{S}$ terms drawn from citations. Databases used are HEPTH (T), HEPEX (E), and CMPLING (M).

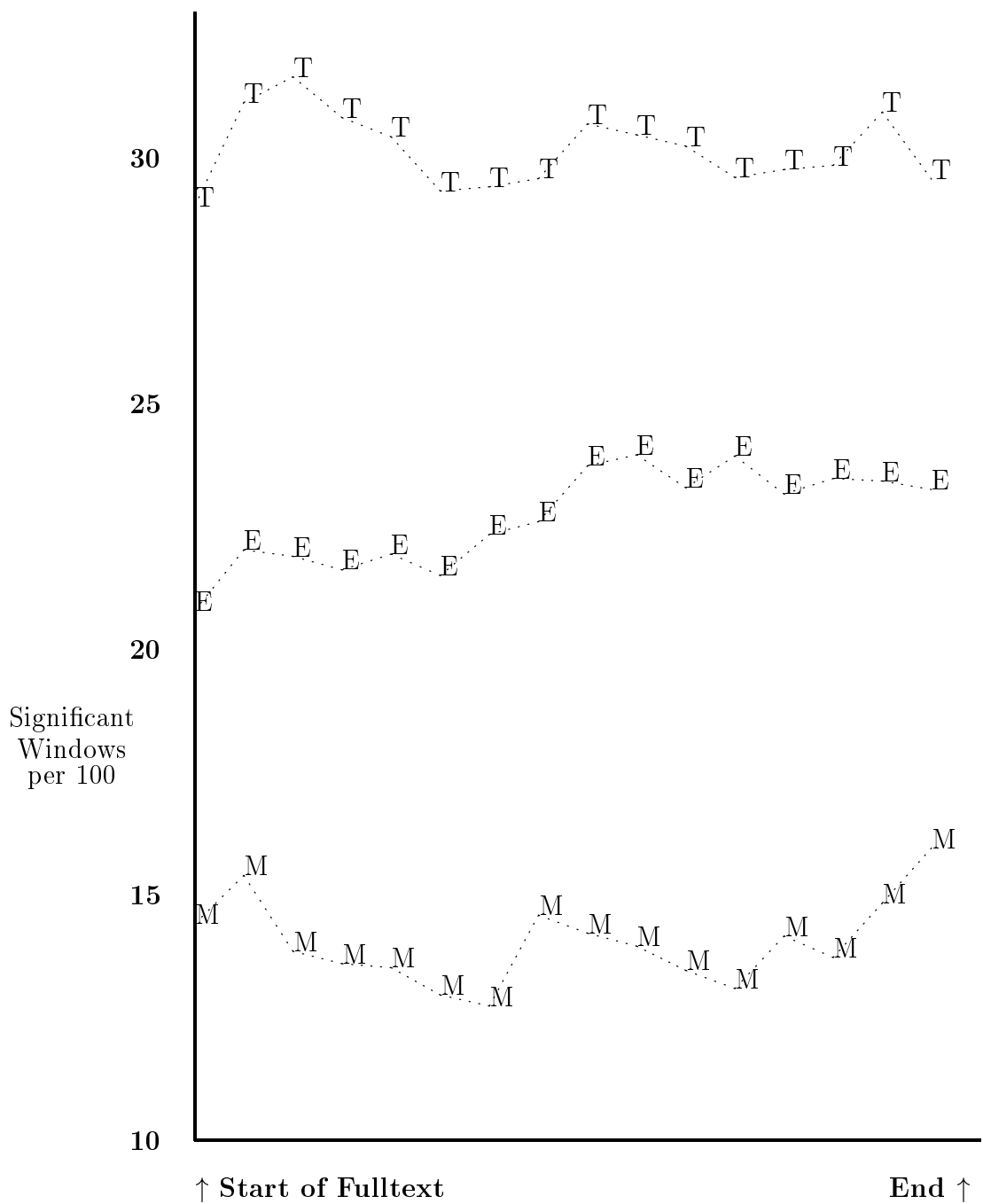

Figure 5: The distribution of significant windows with $\mathcal{S}$ terms drawn from section headers. Databases used are HEPTH (T), HEPEX (E), and CMPLING (M).

The data in Table 2 suggest that the $\mathcal{Q}$ values (as well as SC and WWD) are higher for titles in theoretical disciplines than in the more experimental disciplines. This also holds for citation based $\mathcal{S}$ terms but this does not hold for the abstract field.

The data in Figure 5 represent window rates for $\mathcal{S}$ terms coming from section headers for each document. Interestingly, the windows from HEPTH occur at a clearly higher rate than those of HEPEX (and CMPLING). The data in Table 1 suggest a reason for this. A comparison of HEPTH and HEPEX fields shows similar frequencies in all fields except for section headers, where HEPTH has twice the average number of tokens in a section header than are found in an HEPEX section header. The number of $\mathcal{SW}$s per 100 is about 40% higher for HEPTH than for HEPEX in Figure 5. If one factors in the number of term tokens in the section header field when determining the rate of occurrence of $\mathcal{SW}$s for $\mathcal{S}$ terms based on section headers, it would result in the HEPEX database having a higher rate than HEPTH.

The preceding sections examined how $\mathcal{SW}$s vary in the rate of occurrences depending on their location in documents, the locations in documents from which $\mathcal{S}$ terms are derived, and the disciplinary norms and constraints experienced by an author while the document was produced. We will now turn to more quantitative measures of the relationships between different $\mathcal{S}$ systems across disciplines.

## 8 Overlap in $\mathcal{S}$ Systems

Some sets of $\mathcal{S}$ terms, such as terms in abstracts, may be better than other sets of $\mathcal{S}$ terms at forming those $\mathcal{SW}$s that are likely to be found using other sets of $\mathcal{S}$ terms. These more *precise* sets of $\mathcal{S}$ terms might be seen as including the key terms and phrase constituents in a discipline. The set of $\mathcal{S}$ terms $j$ is a better predictor of the $\mathcal{SW}$s in the intersection of the two sets than is set of $\mathcal{S}$ terms $k$ when the proportion of $\mathcal{SW}$s in $j$ that are also in $k$ is greater than the proportion of $\mathcal{SW}$s in $k$ that are also in $j$.

As an example, consider the case where all the $\mathcal{SW}$s produced by $\mathcal{S}$ terms $j$ are also produced by using $\mathcal{S}$ terms $k$, but not all the $\mathcal{SW}$s in $k$ are in $j$, making $\Pr(k|j) > \Pr(j|k)$. Assume that all the terms in each set of $\mathcal{S}$ terms are of the same grammatical depth, such as might be the case if they were all nouns. We might refer to $j$ as having a higher density of $\mathcal{S}$ terms, or we might want to borrow the concept of *precision* from information retrieval performance and say that $j$ has higher precision than does $k$. Precision may be measured in this application as the percentage of windows produced by an $\mathcal{S}$ system that also are produced by a second $\mathcal{S}$ system, to which it is being compared. If $j$ is more precise than all other sets, $k, l, m, \ldots, z$ for a database, we may say that $j$ is a *most precise predictor* (MPP). The terms in a MPP $j$ may be considered to be those terms *central* to the discipline, whereas those terms in other $\mathcal{S}$ sets $k, l, m, \ldots, z$ are less central and are terms likely to be picked up by fewer $\mathcal{S}$ fields.

Table 4: The probability that a significant window occurs at a particular location for the $\mathcal{S}$ type for the row, given that a significant window is found at the same location with $\mathcal{S}$ terms being from the field indicated by the column header. The window size is $\omega = 3$. Columns with a plus have values that exceed the row value in all cases and are MPPs, while rows with a plus have row values that exceed the column value. A minus sign indicates similarly low valued rows and columns.

|  | | | CF | | |
| --- | --- | --- | --- | --- | --- |
| $\Pr(row\|col)$ | $BI^+$ | $IDF$ | $Title$ | $Abstract$ | $Dict.^-$ |
| $BI^-$ | 1 | .04 | .16 | .19 | .10 |
| $IDF$ | .05 | 1 | .10 | .12 | .29 |
| $Title$ | .18 | .10 | 1 | .34 | .16 |
| $Abstract$ | .24 | .13 | .37 | 1 | .18 |
| $Dict.^+$ | .15 | .36 | .20 | .21 | 1 |

|  | | | HEPEX | | | |
| --- | --- | --- | --- | --- | --- | --- |
| $\Pr(row\|col)$ | $BI$ | $IDF$ | $Title$ | $Abstract^+$ | $Cites$ | $Sec.Hdr.^-$ |
| $BI$ | 1 | .04 | .11 | .19 | .12 | .16 |
| $IDF$ | .03 | 1 | .08 | .07 | .06 | .09 |
| $Title$ | .09 | .10 | 1 | .36 | .21 | .23 |
| $Abstract^-$ | .12 | .06 | .25 | 1 | .17 | .15 |
| $Cites$ | .09 | .07 | .19 | .22 | 1 | .15 |
| $Sec.Hdr.^+$ | .29 | .22 | .46 | .42 | .34 | 1 |

|  | | | HEPTH | | | |
| --- | --- | --- | --- | --- | --- | --- |
| $\Pr(row\|col)$ | $BI$ | $IDF$ | $Title$ | $Abstract^+$ | $Cites$ | $Sec.Hdr.^-$ |
| $BI$ | 1 | .01 | .15 | .17 | .13 | .15 |
| $IDF$ | .04 | 1 | .15 | .15 | .08 | .12 |
| $Title$ | .10 | .16 | 1 | .32 | .25 | .21 |
| $Abstract^-$ | .08 | .13 | .26 | 1 | .18 | .15 |
| $Cites$ | .08 | .09 | .25 | .23 | 1 | .16 |
| $Sec.Hdr.^+$ | .25 | .31 | .51 | .47 | .39 | 1 |

|  | | | CMPLING | | | |
| --- | --- | --- | --- | --- | --- | --- |
| $\Pr(row\|col)$ | $BI^-$ | $IDF$ | $Title$ | $Abstract^+$ | $Cites$ | $Sec.Hdr.$ |
| $BI^+$ | 1 | .10 | .15 | .32 | .22 | .15 |
| $IDF$ | .06 | 1 | .10 | .05 | .06 | .08 |
| $Title$ | .10 | .11 | 1 | .32 | .30 | .31 |
| $Abstract^-$ | .08 | .04 | .24 | 1 | .32 | .31 |
| $Cites$ | .07 | .06 | .26 | .35 | 1 | .31 |
| $Sec.Hdr.$ | .07 | .07 | .26 | .35 | .32 | 1 |

In practice, many common terms, such as *the,* occur in most sets. Subject bearing terms that occur in all the sets are probably good $\mathcal{S}$ terms, while those that occur in few of the sets are less likely to be sublanguage terms.

Consider two sets of $\mathcal{SW}$s, where $j$ are $\mathcal{SW}$s in document titles and $k$ the $\mathcal{SW}$s derived from $\mathcal{S}$ terms in a subject dictionary. Obviously, non-content bearing terms such as *the* and *and* might appear in a document title but would not be in a subject dictionary. There will be a much larger number of terms that appear in the subject dictionary that are not in any titles in a sample database, making $\Pr(dict|title) > \Pr(title|dict)$, suggesting that, in this case, terms in document titles produce a more precise $\mathcal{SW}$ set. We are not saying in this instance that the terms in the subject dictionary are not representative of that discipline, broadly construed, but instead we suggest that for the database in question, $j$ is more precise. If we had a database that covered every topic in the dictionary, then it would be the case that neither $j$ nor $k$ would be more precise than the other.

The probabilities that windows produced in the fulltext using a particular $\mathcal{S}$ method produce the windows generated from another $\mathcal{S}$ method in the fulltext are shown in Table 4. When the first line in Table 4 shows .16 in the row for BI and the column for titles, this indicates that if a window is a significant title window, then there is a 16% chance that the same window is an $\mathcal{SW}$ in the BI $\mathcal{S}$ system.

Some $\mathcal{S}$ systems are likely to be better predictors than others of the windows produced by other $\mathcal{S}$ systems, while some other $\mathcal{S}$ term based systems are easier to predict. Fields with higher values for their rows than for other rows in Table 4 represent those areas that are easiest to predict, no matter what the columns are. These high valued rows (marked with a + in Table 4), represent more general fields that contain less unique information, since they overlap with the terms produced by several other $\mathcal{S}$ systems. Conversely, more precise fields produce columns that have higher values than other fields and represent fields that, as conditioning fields, increase the probability of overlap with other fields. High valued columns marked with a + are MPPs and are good predictors of other fields. For example, Table 4 shows lower values in several databases for the columns for section header derived $\mathcal{S}$ terms and higher values in the columns for abstract derived $\mathcal{S}$ terms.

The information in Table 4 is graphically summarized in Figure 6 through the use of arrows representing the asymmetrical probabilistic relationships between the $\mathcal{S}$ terms from the fields indicated. It shows that $\mathcal{SW}$s produced from $\mathcal{S}$ terms in abstracts are more likely to be found in sets of $\mathcal{SW}$s produced by other fields than if the two fields were reversed. Abstracts may be said to yield more precise $\mathcal{SW}$s than other fields. On the other hand, we find that the fields BI and section headers are generally easiest to predict, that is, they are the least precise. For the CF database, the dictionary terms are easy to predict from several other fields, and this field is similarly less precise for CF.

A similar computer run to that producing Figure 6 or Table 4 but with all term types on a list of 203 stopwords removed from the $\mathcal{S}$ producing fields shows that even with stopwords removed, dictionary terms are still less precise than most other fields.

Figure 6: An arrow indicates that the conditional probablility of the field at the head of the arrow, given the field at the base of the arrow, is greater than the probability wth the two positions reversed. A single straight arrow represents physics alone (HEPTH and HEPEX). A double arrow represents physics and CMPLING. Arrows pointing in the opposite direction should be understood for all relationships that are not covered by the single or double nature represented by the arrows given. Bent arrows represent the CF database relationships with the dictionary terms.

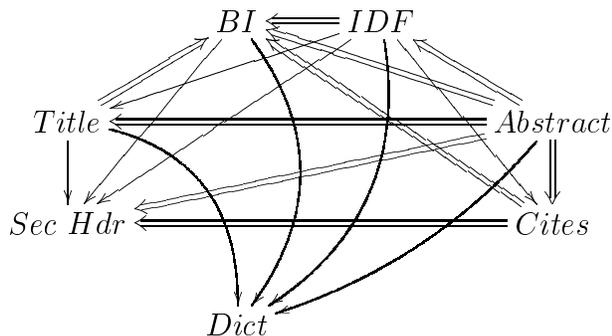

In summary, the set of $\mathcal{S}$ terms derived from abstracts are better predictors than $\mathcal{S}$ terms from other fields of whether a significant window will be produced by the other $\mathcal{S}$ systems. In turn, the significant windows produced by BI and section headers are most likely to be produced by another $\mathcal{S}$ system. We might wish to conclude from this that the $\mathcal{SW}$s derived from abstracts are more useful in getting at other kinds of windows or phrases and that $\mathcal{S}$ terms derived from section headers and BI fields are less useful.

## 9 Type-Token Ratios for Theory and Practice Based Disciplines

The differing rates at which $\mathcal{S}$ terms appear in windows suggest that a more formal examination of the relationship between type and token frequencies may be profitable. A term type is the term itself, while a token is an occurrence of the term. There are always as many tokens as there are types, and almost always many more. The ratio of term types to tokens, the type-token ratio, is inversely related to the degree of reuse of terms. Thus, fields described below with low type-token ratios exhibit higher reuse of terms than fields with higher type-token ratios.

The results reported in earlier sections suggest that more applied fields, such as CF and HEPEX, might be identified by the less frequent reuse of terms, resulting in fewer tokens per term used by an author. Figure 7 shows how the types vary with tokens for the CF and CMPLING databases. For a given number of term types (the horizontal axis), we find that the CMPLING database has a greater number of tokens than does the CF database. The lines in the Figure are produced by graphing the regression curve (best fit curve), allowing a third order polynomial to represent the data. We note that there is not a strong curvature to the data, with the variation from type-token linearity appearing primarily at the lowest term frequencies for the four databases.

Figure 7: The relationships between type and term frequencies for documents from CF and CMPLING databases.

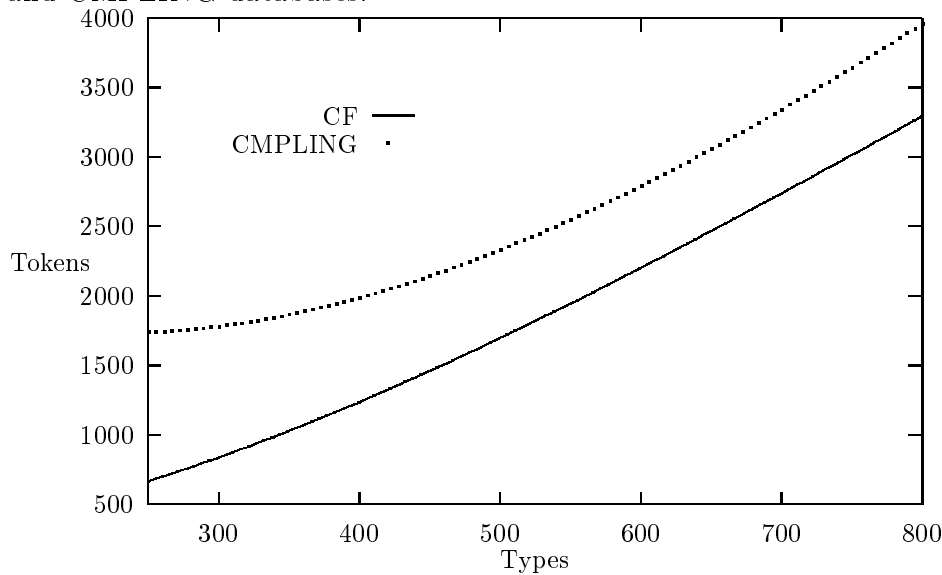

Tagliacozzo (1976) studied the differences between documents of different scientific depth and presented some results suggesting that "highly technical" writings have lower type-token ratios than writings of "lower technicality." These highly technical writings may be more focused and thus produce more reuse of terms than is found in documents with a broader scope.

We have performed a similar test with the type-token ratios for the fields, with the values CF, .29, HEPEX, .27, HEPTH, .23, CMPLING .20. While the type-token ratio should vary somewhat with document length, it is an effective measure of the literary style for different disciplines. Using an ANOVA, the mean type-token values were shown to be significantly different at the .01 level.

It might be the case that the differences in the type-token ratios are due to the number of tokens in the fulltext for each of the databases (Table 1). The data in Figure 7 more clearly captures the relative differences between different number of tokens per type, controlling for length. This data supports our hypothesis. Figure 7 clearly shows the separation between these two theory and practice based disciplines, assuming that the CF medical database is more practice-oriented than is the CMPLING database. The other two databases (not shown) are less distinct, although the difference between theory and practice based documents is still clear.

Limiting the results to the first 500 terms in the fulltext for all documents produces a type-token ratio of CF, .48, HEPEX, .47, HEPTH, .41, and CMPLING, .44. These results are somewhat similar to those obtained with the type-token ratio for the entire documents, with the exception that HEPTH had a lower ratio than did CMPLING. This also controls for length and allows us to examine portions of different sublanguage texts that are likely to be similarly structured, with fewer examples or formulae such as would be found further into an article.

Tagliacozzo "excluded the function words fearing that their high rate of repetitiveness may in some cases obscure differences in type-token ratio of what [were] considered the most significant part of the vocabulary, i.e. the content words." In our work, all the terms were included, and we were still able to distinguish between the type-token ratios, suggesting that Tagliacozzo's concern, while possibly a problem for smaller sets of text abstracts, is not a significant problem for the analysis of fulltext.

It is unclear whether type-token ratios for subfields are significant. For example, examining the type and token frequencies in Table 1 shows that the IDF type-token frequencies are ordered: HEPEX, CF, HEPTH, CMPLING. Other fields, such as titles, show a similar ability to discriminate between our theory and practice documents. In summary, the type-token ratio may be effective when used to separate documents having a more theoretical or a more practical orientation.

## 10 Grammatical Characteristics of Significant Windows

A variety of grammatical constructs are found in windows of differing sizes and in different disciplines. We may compute the probability that a window that is significant (based on term occurrences in a set of $\mathcal{S}$ terms) given that it has grammatical construct $\mathcal{G}$, as $\Pr(\mathcal{SW}|\mathcal{G})$, with the probability that the grammatical construct (if found in a window found not to be statistically significant, $\overline{\mathcal{SW}}$) is $\Pr(\overline{\mathcal{SW}}|\mathcal{G})$. Use of these probabilities may help us decide whether we have an $\mathcal{SW}$ or phrase when we find a particular grammatical construct.

While knowledge of $\Pr(\mathcal{SW}|\mathcal{G})$ is useful when estimating how likely it is that a window is significant given that we find a particular grammatical sequence, it is of limited assistance when used alone in determining how useful a particular grammatical construct is in discriminating between statistically significant and non-significant windows. We may instead use the weight of evidence ($\mathcal{M}_\mathcal{G}$) provided by knowledge of the grammatical construct, that is, the contribution to the our knowledge about the odds that a window is significant that is provided by a grammatical construct. This weight is computed as the log odds that a window is significant, given a grammatical construct, minus the log odds that a window is unconditionally significant (Berger & Wolpert, 1984; Edwards, 1972; Good, 1950; Osteyee & Good, 1974), that is:

$$\mathcal{M}_\mathcal{G} = \log\left(\frac{\Pr(\mathcal{SW}|\mathcal{G})}{\Pr(\overline{\mathcal{SW}}|\mathcal{G})}\right) - \log\left(\frac{\Pr(\mathcal{SW})}{\Pr(\overline{\mathcal{SW}})}\right). \qquad (5)$$

All logarithms were taken to the base 2 for the results presented below.

In Table 5 we have the grammatical constructs in the CF database that have the highest and lowest discrimination values ($\mathcal{M_G}$). Terms in the medical dictionary were the $\mathcal{S}$ terms, serving as a list of the specialized terms in the medical sublanguage. These can be used to help determine whether a particular window is likely to contain dictionary terms. This would have obvious applications to situations where either a dictionary is being constructed or the sublanguage terms are being extracted for applications such as retrieval or indexing.

In Table 6 we show the discrimination capabilities of phrases in discriminating between practical (applied) work (CF and HEPEX) and less applied and more theoretical work (HEPTH and CMPLING). We compute $\Delta_{TP}$, the ability of the grammatical construct to distinguish between theoretical and practical sublanguages, as the sum of the $\mathcal{M_G}$ for CF and HEPEX and the negation of the weights for HEPTH and CMPLING. Systems attempting to identify $\mathcal{S}$ terms or phrases for use in automatic indexing may wish to consider those terms occurring in the grammatical environments with higher $\mathcal{M_G}$ values as being superior candidates for being $\mathcal{S}$ terms, with those with negative $\mathcal{M_G}$ values being highly unlikely candidates.

The grammatical structures that discriminate between disciplines, or on the theory–practice spectrum, may be useful for retrieval, filtering, and document summarization in these, as well as other, disciplines. For example, we have provided evidence that many complex noun phrases are composed of sublanguage terms. The presence of these phrases indicates that the terms in the phrases are more likely to be subject bearing terms that might prove useful for extracting subject bearing phrases or sentences indicating the subject of the document or that might be useful for indexing purposes. Similarly, these complex phrases might be subject to further statistical analysis (e.g. term dependence) by retrieval and filtering programs in order to more accurately rank documents.

## 11  Implications of Research and Summary

In this research we have described several features characterizing fulltext documents from different disciplines and have suggested how documents may be separated based on their disciplines as well as their place on a theory to practice spectrum. This work assumes that windows of interest, $\mathcal{SW}$s, are those windows that are composed of the types of terms that are statistically unlikely to co-occur as they do in the $\mathcal{SW}$s. The data described here was collected from four databases containing, in total, hundreds of documents and hundreds of thousands of $\mathcal{SW}$s. The evidence characterizing different disciplines is based on a few disciplines, and future work will need to examine other disciplines to verify or refute the disciplinary trends proposed here.

A general model of term occurrences in the different disciplines is suggested by the data above. It is clear that some disciplines, which we choose to call "theoretical," have lower type-token ratios, with terms being reused more than in "experimental" disciplines with higher type-token ratios. The theoretical disciplines, with more term reuse, produce less rich fields such as abstracts. This results in less overlap between the abstracts and the fulltext and in lower rates of $\mathcal{SW}$ occurrence. What we call experimental disciplines have fewer tokens per type, making abstracts richer. Terms from abstracts thus have a larger intersection with the terms in fulltext, producing higher rates of $\mathcal{SW}$ occurrence.

Table 5: The sets of part-of-speech tags for the most positivly discriminatory and most negativly discrminatory grammatical constructs for windows of length 1 to 4.

| Grammatical Structures ($\mathcal{G}$) | N | $\Pr(\mathcal{SW}|\mathcal{G})$ | $\mathcal{M}_\mathcal{G}$ |
|---|---|---|---|
| frgn-word | 530 | 0.9132 | 5.1419 |
| noun | 57788 | 0.7354 | 3.2211 |
| adj | 27693 | 0.5522 | 2.0489 |
| proper-noun | 16343 | 0.4741 | 1.5969 |
| interject | 41 | 0.2683 | 0.2991 |
| predet | 44 | 0.2045 | -0.2128 |
| verb | 4779 | 0.1808 | -0.4333 |
| number | 20914 | 0.0028 | -6.7436 |
| modal | 1679 | 0.0012 | -7.9650 |
| prep | 40252 | 0.0008 | -8.5046 |
| frgn-word pl-noun | 143 | 0.9790 | 9.1906 |
| noun noun | 9661 | 0.6119 | 4.3035 |
| adj noun | 13254 | 0.5693 | 4.0486 |
| compartv-adj adj | 44 | 0.5682 | 4.0423 |
| adj adj | 2414 | 0.3964 | 3.0399 |
| proper-noun noun | 2329 | 0.3963 | 3.0391 |
| noun proper-noun | 1024 | 0.3701 | 2.8792 |
| modal verb | 1415 | 0.0021 | -5.2322 |
| number comma | 1012 | 0.0020 | -5.3337 |
| pst-tns-verb vrb-past-prt | 3075 | 0.0016 | -5.6157 |
| proper-noun frgn-word pl-noun | 133 | 1.0000 | 22.8575 |
| frgn-word pl-noun period | 95 | 1.0000 | 22.3721 |
| compartv-adj adj noun | 29 | 1.0000 | 20.6602 |
| adj noun noun | 1871 | 0.8386 | 4.8917 |
| vrb-past-prt noun noun | 200 | 0.8250 | 4.7516 |
| noun prep frgn-word | 22 | 0.8182 | 4.6844 |
| adj adj noun | 1361 | 0.7957 | 4.4764 |
| noun prep det | 4409 | 0.0016 | -6.7821 |
| prep det noun | 5782 | 0.0010 | -7.3964 |
| det noun prep | 4646 | 0.0009 | -7.6660 |
| noun to noun noun | 24 | 0.9167 | 7.0635 |
| vrb-past-prt noun noun noun | 20 | 0.8000 | 5.6041 |
| adj noun noun noun | 236 | 0.7458 | 5.1566 |
| noun conj adj noun | 231 | 0.7446 | 5.1477 |
| adj adj noun noun | 178 | 0.7247 | 5.0006 |
| noun adj noun noun | 155 | 0.7032 | 4.8487 |
| noun conj noun noun | 175 | 0.6971 | 4.8069 |
| noun pst-tns-verb vrb-past-prt prep | 690 | 0.0043 | -4.2351 |
| adj pl-noun prep det | 491 | 0.0041 | -4.3295 |
| vrb-past-prt prep det noun | 992 | 0.0020 | -5.3471 |

Table 6: Grammatical constructs can be used to distinguish between theoretical (HEPTH and CMPLING) and more practical documents (CF and HEPEX). $\Delta_{TP}$ represents the sum of the $\mathcal{M}_\mathcal{G}$ values for the experimental disciplines and the negation of the values for the theoretical disciplines.

| Grammatical Structures ($\mathcal{G}$) | $\mathcal{M}_\mathcal{G}$ Values for | | | | |
| --- | --- | --- | --- | --- | --- |
| | CF | HEPEX | HEPTH | CMPLING | $\Delta_{TP}$ |
| noun noun noun noun | 1.28 | 0.28 | 1.25 | -3.71 | 4.02 |
| noun adj noun noun | 1.40 | 1.22 | 0.35 | -1.27 | 3.54 |
| adj noun pl-noun conj | 1.02 | 1.44 | -0.82 | -0.25 | 3.53 |
| prep adj pl-noun conj | 2.10 | 0.67 | -0.98 | 0.37 | 3.38 |
| adj noun noun pl-noun | 1.62 | 1.32 | -0.69 | 0.41 | 3.22 |
| prep det number pl-noun | 1.01 | 1.00 | -0.53 | -0.65 | 3.19 |
| det adj pl-noun conj | 2.01 | 1.34 | -0.30 | 0.69 | 2.96 |
| comma det adj pl-noun | 1.52 | 1.57 | 0.01 | 0.16 | 2.92 |
| adj noun noun noun | 1.24 | 0.93 | 0.21 | -0.90 | 2.86 |
| adv verb det noun | -0.86 | -1.14 | -2.77 | -2.09 | 2.85 |
| prep adj noun noun | 1.19 | 1.10 | 0.00 | -0.50 | 2.80 |
| pl-noun prep proper-noun noun | 2.21 | 1.52 | 1.35 | -0.39 | 2.77 |
| noun conj adj noun | 1.44 | 0.14 | -0.93 | -0.22 | 2.74 |
| 3rd-pers-vrb det noun prep | -0.58 | 0.24 | 1.52 | 1.45 | -3.32 |
| det noun to det | 1.41 | 1.69 | 3.48 | 3.07 | -3.45 |
| to verb prep det | -0.17 | -0.06 | 1.46 | 1.86 | -3.55 |
| prep adj pl-noun 3rd-pers-vrb | -0.25 | -0.51 | 0.35 | 2.73 | -3.84 |
| noun prep det number | -0.97 | -1.08 | 1.54 | 0.34 | -3.94 |
| adj noun 3rd-pers-vrb vrb-past-prt | -2.33 | -2.66 | -0.53 | -0.51 | -3.95 |
| noun noun 3rd-pers-vrb vrb-past-prt | -1.44 | -2.08 | 0.51 | 0.00 | -4.03 |
| prep wh-det det noun | -0.33 | 0.23 | 2.14 | 1.96 | -4.20 |
| det noun prep number | 0.17 | -3.40 | 0.97 | 0.06 | -4.26 |
| comma adv comma det | 1.67 | 0.05 | 1.35 | 4.75 | -4.37 |
| prep number prep det | 0.90 | -1.33 | 2.55 | 2.44 | -5.43 |
| number noun prep det | -1.54 | -1.16 | 1.15 | 1.65 | -5.50 |

The *fulltext sublanguage differentiation* hypothesis suggests that there are significant differences between the vocabulary, grammatical, and style in fulltext documents from different disciplines. We have provided evidence supporting this hypothesis. For example, Figures 3 through 5 show that individual disciplines may be characterized by the varying rate of $\mathcal{SW}$ occurrences as one moves through the fulltext of documents. Similarly, the data in Table 6 show some of the variation in syntactic structure in different academic disciplines.

A primary focus of this study is on describing the characteristics of different features of documents in different disciplines. This information is derived in such a way that it may be used in inferential environments, such as an automatic classification system. One useful clue in identifying the disciplinary source of a document is the "shape of the curve" in Figures 3 through 5 above, that is, the change in the rate of $\mathcal{SW}$ occurrences as one moves through the fulltext of a document. The differences between $\mathcal{SW}$s derived from different sets of $\mathcal{S}$ terms and different document fields are significant and useful.

Additionally, the density of $\mathcal{S}$ terms in $\mathcal{SW}$s, or within window density (WWD), and the rate at which $\mathcal{S}$ terms occur in $\mathcal{SW}$s, the $\mathcal{S}$ concentration (SC), are useful at distinguishing between sublanguages. Using these measures, we have suggested a composite measure, $\mathcal{Q}$, that allows us to examine the overall quality of a windowing system. In addition, the type-token ratios may be used to discriminate between different sublanguages. A statistical analysis showed that the mean type-token ratios were significantly different for the different disciplines.

These techniques may also be used to help distinguish between documents at various points on a theory-practice spectrum. The regression line for the type-token ratios for the CF and CMPLING databases, for example, show how the type-token ratios may be used to discriminate between disciplines emphasizing theory verses practice, between articles emphasizing understanding and those emphasizing applications. The type-token ratios for similarly sized documents may prove useful in automatic classification, with similarly sized documents with type-token ratios in the low .20's being more likely theoretical documents and those in the high .20's being experimental or applied documents.

The study of statistically significant windows may be used to isolate more meaningful terms and concepts in a document and in a discipline. Knowledge of the $\mathcal{SW}$s and their location, number, and characteristics may be used in automatic indexing, dictionary construction, and the automatic classification of documents. Future work may be able to analyze sublanguage terms and $\mathcal{SW}$s through the appropriate parameterization of the Zipf distribution. We hope to continue the study of syntactic groupings for the automatic identification of key terms and phrases in documents, as well as to isolate those subject bearing parts of documents for use in systems that summarize documents. Additionally, allowing us to describe more important parts of documents will allow us to focus the analytic study of retrieval and filtering system performance (Losee, 1988, 1995a, 1996a) on those parameters likely to have the greatest impact on performance.